\documentclass[10pt]{article}
\usepackage{colacl}
\usepackage{epsfig}
\title{Indexing with WordNet synsets can improve text retrieval}

\author{Julio Gonzalo \and Felisa Verdejo \and Irina Chugur \and Juan Cigarr\'an
\\ UNED \\
  Ciudad Universitaria, s.n. \\
  28040 Madrid - Spain \\
  {\tt \{julio,felisa,irina,juanci\}@ieec.uned.es}}

\begin{document}

\maketitle
\begin{abstract}
The classical, vector space model for text retrieval is shown to give
better results (up to 29\% better in our experiments) if WordNet
synsets are chosen as the indexing space, instead of word forms. This
result is obtained for a manually disambiguated test collection (of
queries and documents) derived from the {\sc Semcor} semantic
concordance. The sensitivity of retrieval performance to (automatic)
disambiguation errors when indexing documents is also
measured. Finally, it is observed that if queries are not
disambiguated, indexing by synsets performs (at best) only as good as 
standard word indexing. 
\end{abstract}


\section{Introduction}
\label{sec:previousexp}

Text retrieval deals with the problem of
finding all the relevant documents in a text collection for a given 
user's query. A large-scale semantic database such as WordNet
\cite{Miller-90b} 
seems to have a great potential for this task. There are, at least,
two obvious reasons: 

\begin{itemize}
\item It offers the possibility to discriminate word senses in documents and
  queries. This would prevent matching {\em spring} in its ``metal
  device'' sense with documents mentioning {\em spring} in the sense of
  {\em springtime}. And then retrieval accuracy could be improved.
\item WordNet provides the chance of matching semantically related words. For
  instance, {\em spring}, {\em fountain}, {\em outflow}, {\em
    outpouring}, in the appropriate senses, can be identified as
  occurrences of the same concept, `{\em natural flow of ground
    water}'. And beyond synonymy, WordNet can be used to measure semantic
  distance between occurring terms to get more sophisticated ways of
  comparing documents and queries.
\end{itemize}

However, the general feeling within the information retrieval
community is that dealing explicitly with semantic information 
does not improve significantly the performance of text retrieval
systems. This impression is founded on the results of
some experiments measuring the role of Word Sense Disambiguation (WSD) 
for text retrieval, on one hand, and some attempts to exploit the
features of WordNet and other lexical databases, on the other hand.

In \cite{Sanderson-94}, word sense ambiguity is shown to produce
only minor effects on retrieval accuracy, apparently confirming that
query/document matching strategies already perform an implicit
disambiguation. Sanderson also estimates that if explicit WSD is
performed with less than 90\% accuracy, the results 
are worse than non disambiguating at all. In his experimental setup,
ambiguity is introduced artificially in the documents, substituting
randomly chosen pairs of words (for instance, {\em banana} and {\em
  kalashnikov}) with artificially ambiguous terms ({\em
  banana/kalashnikov}). While his results are very interesting, it
remains unclear, in our opinion, whether they would be corroborated
with real occurrences of ambiguous words. There is also other minor
weakness in Sanderson's experiments. When he ``disambiguates'' a term
such as {\em spring/bank} to get, for instance, {\em bank}, he has
done only a partial disambiguation, as {\em bank} can be used in more
than one sense in the text collection.

Besides disambiguation, many attempts have been done to exploit
WordNet for text retrieval purposes. Mainly two aspects have been
addressed: the enrichment of queries with semantically-related
terms, on one hand, and the comparison of queries and documents via
conceptual distance measures, on the other. 

Query expansion with WordNet has shown to be potentially relevant to
enhance recall, as it permits matching relevant
documents that could not contain any of the query terms
\cite{Smeaton-95}. However, it has produced few successful
experiments.  
For instance, \cite{Voorhees-94} 
manually expanded 50 queries over  
a TREC-1 collection \cite{Harman-93} using synonymy and other semantic relations
from WordNet 1.3. Voorhees found that the expansion was
useful with  short, incomplete queries, and rather useless for
complete topic statements -where other expansion techniques worked
better-. For short queries, it remained the problem of
selecting the expansions automatically; doing it badly could
degrade retrieval performance rather than enhancing it. In
\cite{Richardson-95}, a combination of rather sophisticated techniques 
based on WordNet, including automatic disambiguation and measures of
semantic relatedness between query/document concepts resulted in a
drop of effectiveness. Unfortunately, the effects of WSD errors could
not be discerned from the accuracy of the retrieval strategy. 
However, in \cite{Smeaton-96}, retrieval
on a small collection of image captions - that is, on very short
documents - is reasonably improved using measures of conceptual
distance between words based on WordNet 1.4. Previously, captions
and queries had been manually disambiguated against WordNet. The
reason for such success is that with very short documents (e.g. {\em
  boys playing in the sand}) the chance of finding the original terms
of the query (e.g. of {\em   children running on a beach}) are much lower
than for average-size documents (that typically include many
phrasings for the same concepts). These results are in agreement with
\cite{Voorhees-94}, but it remains the question of whether the
conceptual distance matching would scale up to longer documents and
queries. In addition, the experiments in \cite{Smeaton-96} only
consider nouns, while WordNet offers the chance to use all open-class
words (nouns, verbs, adjectives and adverbs). 

Our essential retrieval strategy in the experiments reported here is
to adapt a classical vector model based system, using WordNet synsets
as indexing space instead of word forms. This approach combines two
benefits for retrieval: one, that terms are fully disambiguated (this
should improve precision); and two, that equivalent terms can be
identified (this should improve recall). Note that query expansion
does not satisfy the first condition, as the terms used to expand are
words and, therefore, are in turn ambiguous. On the other hand, plain
word sense disambiguation does not satisfy the second condition, as
equivalent senses of two different words are not matched. Thus,
indexing by synsets gets maximum matching and minimum spurious
matching, seeming a good starting point to study text retrieval with
WordNet. 

Given this approach, our goal is to test two main issues which are not
clearly answered -to our knowledge- by the experiments mentioned above:

\begin{itemize}
\item Abstracting from the problem of sense disambiguation, what
  potential does WordNet offer for text retrieval? In particular, we
  would like to extend experiments with manually disambiguated
  queries {\em and} documents to average-size texts. 
\item Once the potential of WordNet is known for a manually
  disambiguated collection, we want to test the sensitivity of
  retrieval performance to disambiguation errors introduced by
  automatic WSD. 
\end{itemize}

This paper reports on our first results answering these
questions. 
The next section describes the test collection that we have 
produced. The experiments are described in Section~3, and the last
Section discusses the results obtained.

\section{The test collection}

The best-known publicly available corpus hand-tagged with WordNet senses is
{\sc Semcor} \cite{Miller-93}, a subset of the Brown Corpus of about
100 documents that occupies about 11 Mb. (including tags) The
collection is rather heterogeneous, covering politics, sports, music,
cinema, philosophy, excerpts from fiction novels, scientific
texts... A new, bigger version has been made available recently
\cite{Landes-98}, but we have not still adapted it for our collection.

We have adapted {\sc Semcor} in order to build a test collection -that we
call {\sc IR-Semcor}- in four manual steps:

\begin{table*}
\label{table1}
\begin{center}
\begin{tabular}{|lc|} \hline
{\bf Experiment} & {\bf \begin{tabular}{c} \% correct document \\ retrieved in
first place \end{tabular}} \\ & \\ 
Indexing by synsets &  62.0 \\
Indexing by word senses &  53.2 \\
Indexing by words (basic SMART) &  48.0 \\
\hline 
Indexing by synsets with a 5\% errors ratio & 62.0 \\
Id. with 10\% errors ratio & 60.8 \\
Id. with 20\% errors ratio & 56.1 \\
Id. with 30\% errors ratio & 54.4 \\
Indexing with all possible synsets (no disambiguation) &  52.6 \\
Id. with 60\% errors ratio & 49.1 \\
\hline 
Synset indexing with non-disambiguated queries & 48.5 \\
Word-Sense indexing with non-disambiguated queries & 40.9 \\
\hline
\end{tabular}
\end{center}
\caption{Percentage of correct documents retrieved in first place}
\end{table*}

\begin{itemize}
\item We have split the documents to get coherent chunks of text
  for retrieval. We have obtained 171 fragments that constitute our
  text collection, with an average length of 1331 words per fragment.
\item We have extended the original {\em TOPIC} tags of the Brown
  Corpus with a hierarchy of subtags, assigning a set of tags to each 
  text in our collection. This is not used in the experiments reported 
  here. 
\item We have written a summary for each of the fragments, with lengths
  varying between 4 and 50 words and an average of 22 words per
  summary. Each summary is a human explanation of the text contents,
  not a mere bag of related keywords. 
  These summaries serve as queries on the text collection, and then 
  there is exactly one relevant document per query. 
  \item Finally, we have hand-tagged each of the summaries with WordNet 1.5
  senses. When a word or term was not present in the database, it was left
  unchanged. In general, such terms correspond to groups (vg. {\em
    Fulton\_County\_Grand\_Jury}), persons ({\em Cervantes}) or
  locations ({\em Fulton}). 
\end{itemize}

We also generated a list of ``stop-senses'' and a list of
``stop-synsets'', automatically translating a standard list of stop
words for English.  

Such a test collection offers the chance to measure the adequacy of
WordNet-based approaches to IR independently from the disambiguator
being used, but also offers the chance to measure the role of
automatic disambiguation by introducing different rates of 
``disambiguation errors'' in the collection. The only disadvantage is the
small size of the collection, which does not allow fine-grained
distinctions in the results. However, it has proved large
enough to give meaningful statistics for the experiments reported here.

Although designed for our concrete text retrieval testing purposes,
the resulting database could also be useful for many other tasks. For
instance, it could be used to evaluate automatic summarization systems 
(measuring the semantic relation between the manually written and
hand-tagged summaries of {\sc IR-Semcor} and the output of text
summarization systems) and other related tasks.

\section{The experiments}

We have performed a number of experiments using a standard
vector-model based text retrieval system, {\sc Smart}
\cite{Salton-71},  and three 
different indexing spaces: the original terms in the documents (for standard
{\sc Smart} runs), the word-senses corresponding to the document 
terms (in other words, a manually disambiguated version of the documents) 
and the WordNet synsets corresponding to the document terms (roughly
equivalent to concepts occurring in the documents). 

These are all the experiments considered here:

\begin{figure*}
\epsfig{file=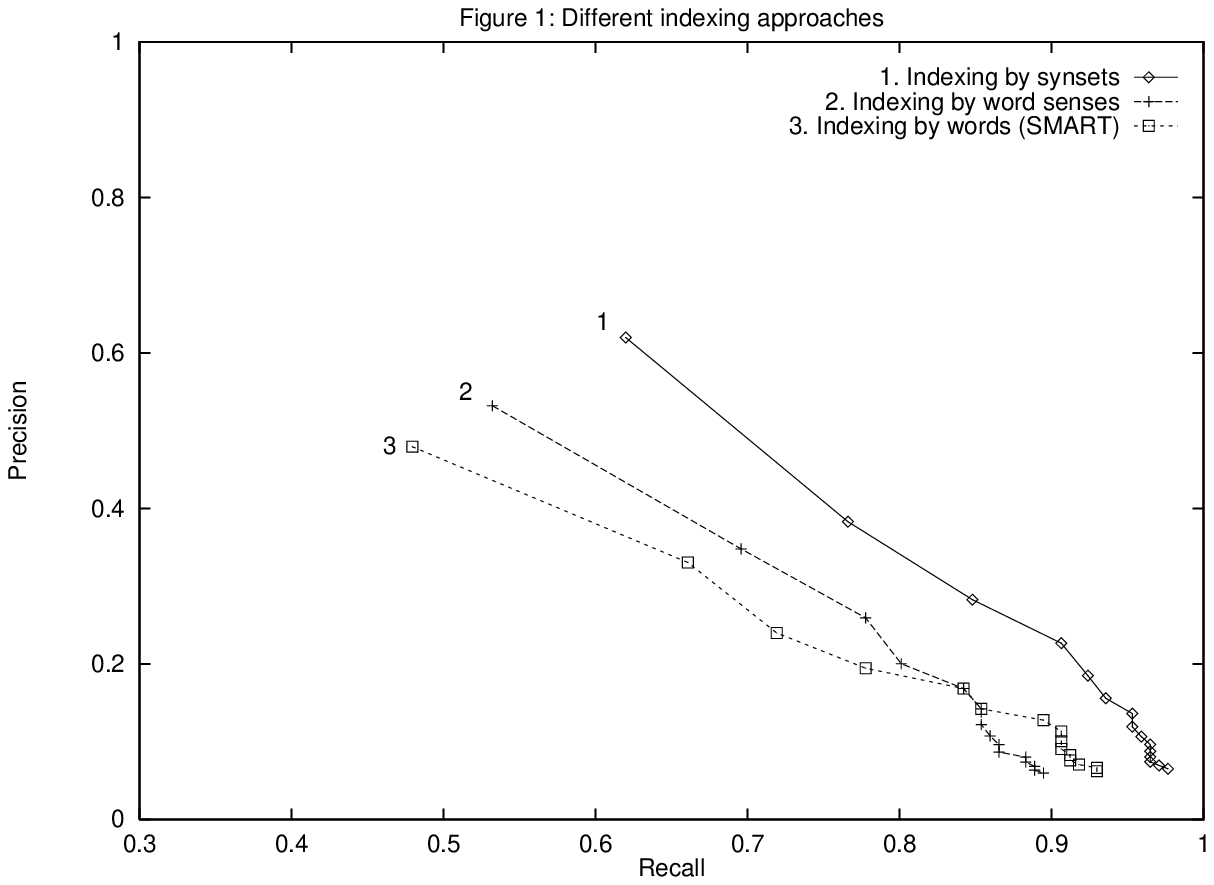, width=15cm, angle=0}
\vfill
\end{figure*}

\begin{enumerate}
\item The original texts as documents and the summaries as
  queries. This is a classic {\sc Smart} run, with the peculiarity that
  there is only one relevant document per query. 
\item Both documents (texts) and queries (summaries) are indexed in
  terms of word-senses. That means that we disambiguate manually all
  terms. For instance ``{\em debate}'' might be substituted with
  ``{\em debate\%1:10:01::}''. The three numbers denote the part of
  speech, the WordNet lexicographer's file and the sense number within
  the file. In this case, it is a noun belonging to the
  {\em noun.communication} file.

  With this collection we can see if plain
  disambiguation is helpful for retrieval, because word senses are
  distinguished but synonymous word senses are not identified.
\item In the previous collection, we substitute each word sense for a
  unique identifier of its associated synset. For instance,
  ``{\em debate\%1:10:01::}'' is substituted with ``$n04616654$'', which is an
  identifier for 
  {\em \begin{center}
      ``\{argument, debate1\}'' (a discussion in
      which   reasons are advanced for and against some proposition or proposal;
      "the argument over foreign aid goes on and on")
    \end{center}}
  This collection
  represents conceptual indexing, as equivalent word senses are
  represented with a unique identifier. 
\item We produced different versions of the synset indexed collection, 
    introducing fixed percentages of erroneous synsets. Thus we simulated a
  word-sense disambiguation process with 5\%, 10\%, 20\%, 30\% and 60\% error
  rates. The errors were introduced randomly in the ambiguous words of
  each document. With this set of experiments we can measure the
  sensitivity of the retrieval process to disambiguation errors.
\item To complement the previous experiment, we also prepared
  collections indexed with all possible meanings (in their word sense
  and synset versions) for each term. This represents a lower bound
  for automatic disambiguation: we should not disambiguate if
  performance is worse than considering all possible senses for every
  word form.
\item We produced also a non-disambiguated version of the queries
  (again, both in its word sense and synset variants). This set of
  queries was run against the manually disambiguated collection.
\end{enumerate}

In all cases, we compared {\tt atc} and {\tt nnn} standard weighting
schemes, and they produced very similar results. Thus we only report
here on the results for {\tt nnn} weighting scheme.

\section{Discussion of results}

\subsection{Indexing approach}

In Figure~1 we compare different indexing approaches: indexing by
synsets, indexing by words (basic SMART) and indexing by word
senses (experiments 1, 2 and 3). The leftmost point in each curve
represents the percentage of documents that were successfully ranked
as the most relevant for its summary/query. The next point represents
the documents retrieved as the first or the second most relevant to
its summary/query, and so 
on. Note that, as there is only one relevant document per query, the
leftmost point is the most representative of each curve. Therefore, we
have included this results separately in Table~1. 

The results are encouraging:

\begin{itemize}
\item  {\bf Indexing by WordNet synsets} produces a remarkable improvement
  on our test collection. A 62\% of the documents are retrieved in
  first place by its summary, against 48\% of the basic {\sc Smart}
  run. This represents 14\% more documents, a 29\% improvement with
  respect to {\sc Smart}. This is an excellent result,
  although we should keep 
  in mind that is obtained with manually disambiguated queries and
  documents. Nevertheless, it shows that WordNet can greatly enhance 
  text retrieval: the problem resides in achieving accurate automatic
  Word Sense Disambiguation.
\item {\bf Indexing by word senses} improves performance when considering 
  up to four documents retrieved for each query/summary, although it is 
  worse than indexing by synsets. This confirms our intuition that
  synset indexing has advantages over plain word sense
  disambiguation, because it permits matching semantically similar
  terms. 

\begin{figure*}
\epsfig{file=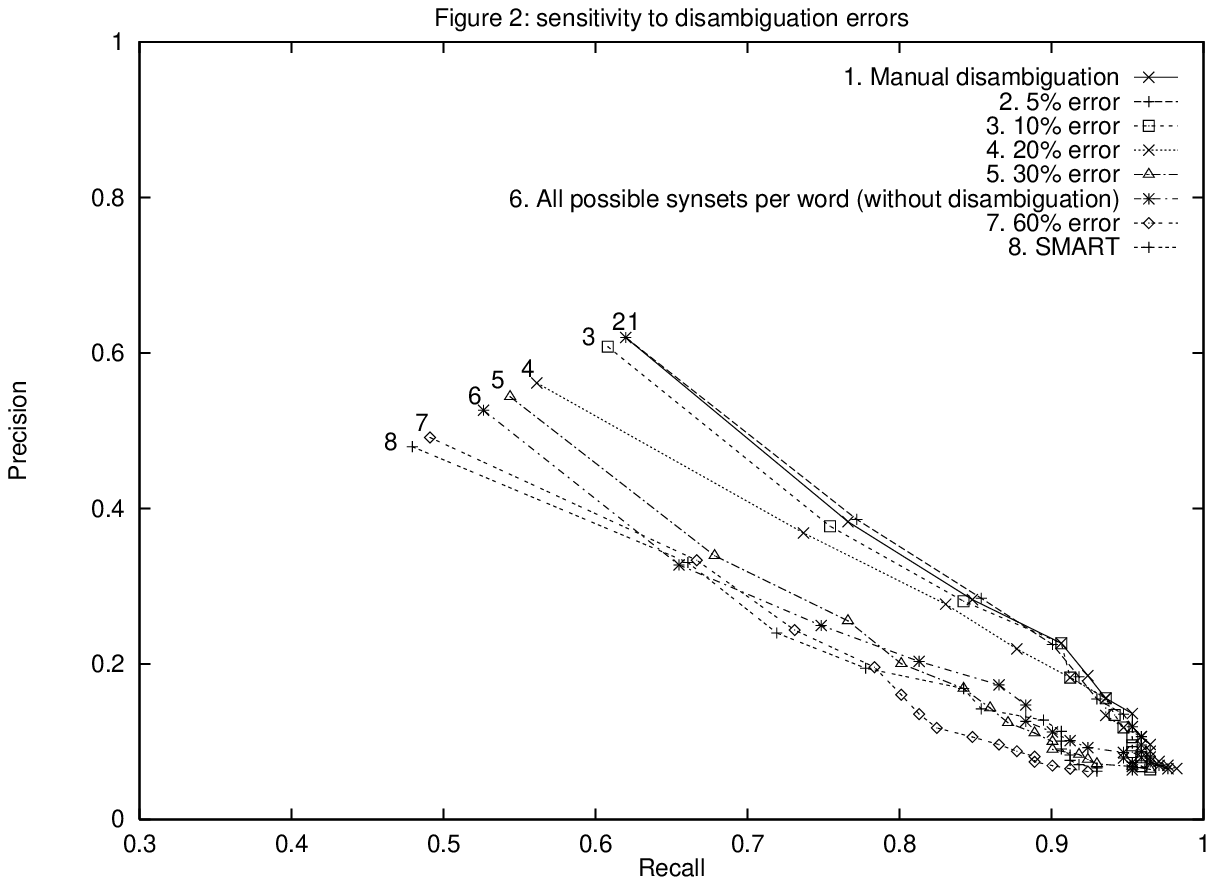, width=15cm, angle=0}
\vfill
\end{figure*}
  
  Taking only the first document retrieved for each summary, the
  disambiguated collection gives a 53.2\% success against a 48\% of
  the plain {\sc Smart} query, which represents a 11\% improvement.
  For recall levels higher than 0.85, however, the disambiguated
  collection performs slightly worse.  This may seem surprising, as word sense
  disambiguation should only increase our knowledge about queries and
  documents. But we should bear in mind that WordNet 1.5 is not the
  perfect database for text retrieval, and indexing by word senses
  prevents some matchings that can be useful for retrieval. For 
  instance, {\em design} is used as a noun repeatedly in one of the documents,
  while its summary uses {\em design} as a verb. WordNet 1.5 does not
  include cross-part-of-speech semantic relations, so this relation
  cannot be used with word senses, while term indexing simply (and
  successfully!) does not distinguish them. Other problems of WordNet
  for text retrieval include too much fine-grained sense-distinctions
  and lack of domain information; see \cite{Gonzalo-98} for a more
  detailed discussion on the adequacy of WordNet structure for text
  retrieval. 
\end{itemize}

\begin{figure*}
\epsfig{file=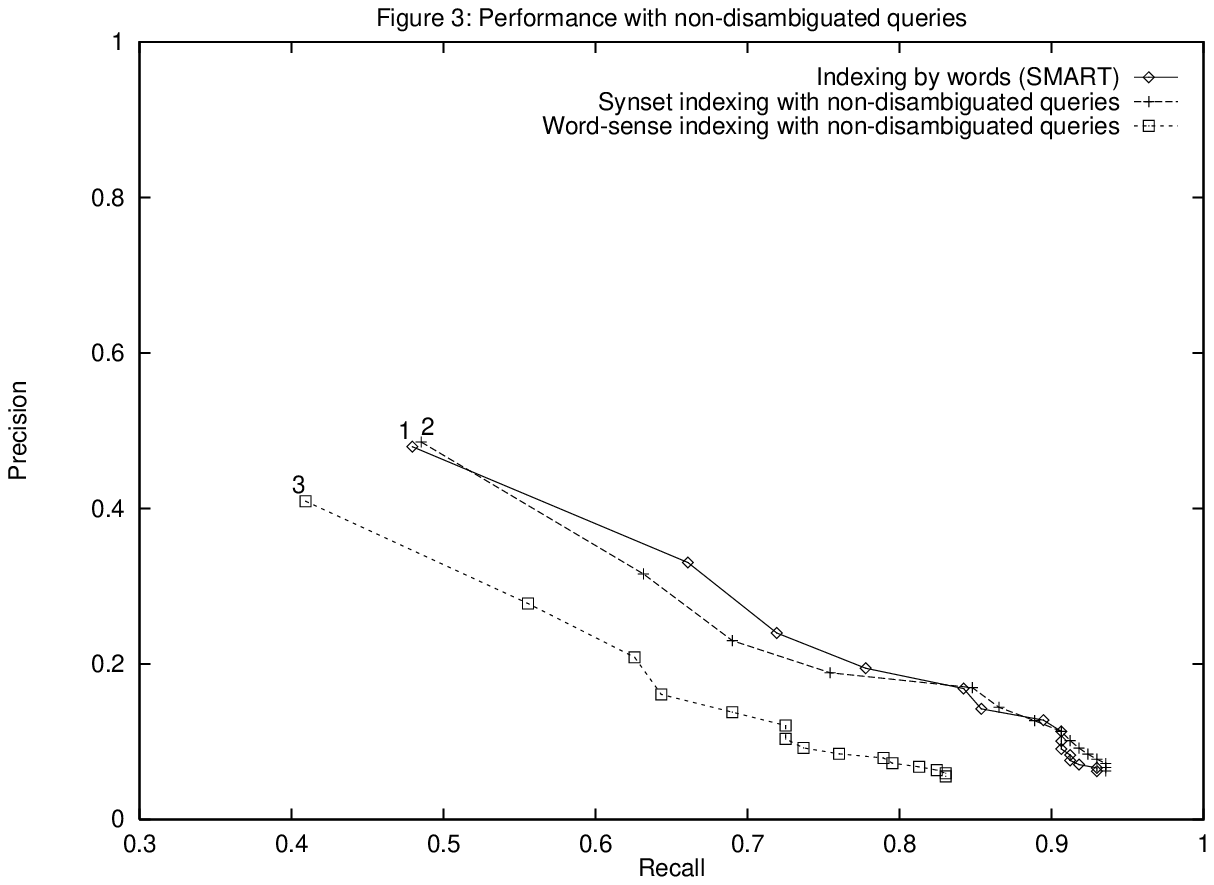, width=15cm, angle=0}
\vfill
\end{figure*}

\subsection{Sensitivity to disambiguation errors}

Figure~2 shows the sensitivity of the synset indexing system to
degradation of disambiguation accuracy (corresponding to the
experiments 4 and 5 described above). From the plot, it can be seen that:

\begin{itemize}
\item Less than 10\% disambiguating errors does not 
  substantially affect performance. This is roughly in agreement
  with \cite{Sanderson-94}.
\item  For error ratios over 10\%, the performance degrades
  quickly. This is also in agreement with \cite{Sanderson-94}.
\item However, indexing by synsets remains better than the basic {\sc
    Smart} run up to 30\% disambiguation errors. From 30\% to 60\%,
  the data does not show significant differences with standard {\sc
    Smart} word indexing. This prediction differs
  from \cite{Sanderson-94} result (namely, that it is better not to
  disambiguate below a 90\% accuracy). The main difference is that we
  are using concepts rather than word senses. But, in addition, it
  must be noted that Sanderson's setup used artificially created
  ambiguous pseudo words (such as {\em `bank/spring'}) which are not
  guaranteed to behave as real ambiguous words. Moreover, what he
  understands as disambiguating is selecting -in the example- {\em
    bank} or {\em spring} which remain to be ambiguous words
  themselves. 
\item If we do not disambiguate, the performance is slightly worse
  than disambiguating with 30\% errors, but remains better than term
  indexing, although the results are not definitive. An interesting
  conclusion is that, if we can disambiguate reliably the queries,
  WordNet synset indexing could improve performance even without
  disambiguating the documents. This could be confirmed on much
  larger collections, as it does not involve manual disambiguation.
\end{itemize}

It is too soon to say if state-of-the-art WSD
techniques can perform with less than 30\% errors, because each
technique is evaluated in fairly different settings. Some of the best
results on a comparable setting (namely, disambiguating against WordNet,
evaluating on a subset of the Brown Corpus, and treating the 191 most
frequently occurring and ambiguous words of English) are reported
reported in \cite{Ng-97}. They reach a 58.7\% accuracy on a Brown
Corpus subset and a 75.2\% on a subset of the Wall Street Journal
Corpus. A more careful evaluation of the role of WSD is needed to know 
if this is good enough for our purposes.

Anyway, we have only emulated a WSD algorithm that just picks up one
sense and 
discards the rest. A more reasonable approach here could be
giving different probabilities for each sense of a word, and use them
to weight synsets in the vectorial representation of documents and
queries. 

\subsection{Performance for non-disambiguated queries}

In Figure~3 we have plot the results of runs with a non-disambiguated
version of the queries, both for word sense indexing and synset
indexing, against the manually disambiguated collection (experiment 6).
The synset run performs approximately as the basic {\sc Smart} run. It seems
therefore useless to apply conceptual indexing if no disambiguation of 
the query is feasible. This is not a major problem in an interactive
system that may help the user to disambiguate his query, but it must
be taken into account if the process is not interactive and the query
is too short to do reliable disambiguation.

\section{Conclusions}

We have experimented with a retrieval approach based on indexing in
terms of WordNet synsets instead of word forms, trying to address two
questions: 1) what potential does WordNet offer for text retrieval,
abstracting from the problem of sense disambiguation, and 2) what is
the sensitivity of retrieval performance to disambiguation errors. The 
answer to the first question is that indexing by synsets can be very
helpful for text retrieval; our experiments give up to a 29\%
improvement over a standard {\sc Smart} run indexing with words. 
We believe that these results have to be further contrasted, but they
strongly suggest that WordNet can be more useful to Text Retrieval
than it was previously thought.

The second question needs further, more fine-grained,
experiences to be clearly answered. However, for
our test collection, we find that error rates below 30\% still produce better
results than standard word indexing, and that from 30\% to 60\%
error rates, it does not behave worse than the standard {\sc Smart}
run. We also find that the queries have to be disambiguated to take
advantage of the approach; otherwise, the best possible results with
synset indexing does not improve the performance of standard word indexing.

Our first goal now is to improve our retrieval system in many
ways, studying how to enrich the query with semantically related synsets,
how to compare documents and queries using semantic information beyond
the cosine measure, and how to obtain weights for synsets according to
their position in the WordNet hierarchy, among other issues.

A second goal is to apply synset indexing in a Cross-Language
environment, using the {\em EuroWordNet} multilingual database
\cite{Gonzalo-98}. Indexing by synsets offers a neat way of performing 
language-independent retrieval, by mapping synsets into 
the EuroWordNet {\em InterLingual Index} that links monolingual
wordnets for all the languages covered by EuroWordNet.

\noindent{\bf Acknowledgments}

{\small
This research is being supported by the European Community, project LE
\#4003 and also partially by the Spanish government, project
TIC-96-1243-CO3-O1. We are indebted to Ren\'ee Pohlmann for giving us good
pointers at an early stage of this work, and to Anselmo Pe\~nas and
David Fern\'andez for their help finishing up the test collection.
}

\bibliographystyle{acl}

\end{document}